\documentclass{article} %
\usepackage{iclr2025_conference,times}

\usepackage{amsmath,amsfonts,bm}

\def\eqref#1{equation~\ref{#1}}

\def\1{\bm{1}}

\DeclareMathAlphabet{\mathsfit}{\encodingdefault}{\sfdefault}{m}{sl}
\SetMathAlphabet{\mathsfit}{bold}{\encodingdefault}{\sfdefault}{bx}{n}

\usepackage[pagebackref=true,breaklinks,colorlinks,citecolor=blue,linkcolor=blue,urlcolor=red,hypertexnames=false]{hyperref}

\usepackage{graphicx}
\usepackage{wrapfig}  %
\usepackage{subcaption}
\usepackage{multirow}
\usepackage{booktabs}
\usepackage{adjustbox}
\usepackage{svg}
\usepackage[tableposition=bottom]{caption}
\usepackage{paralist}
\usepackage{colortbl}
\usepackage{xcolor}

\setdefaultleftmargin{2em}{}{}{}{}{}

\newcommand{\boldparagraph}[1]{\par\vspace{0.2em}\noindent{\bf #1.}}

\newcommand*{\circled}[1]{\lower.7ex\hbox{\tikz\draw (0pt, 0pt)%
    circle (.5em) node {\makebox[1em][c]{\small #1}};}}
\newcommand{\best}[1]{\textbf{#1}}
\newcommand{\second}[1]{\underline{#1}}

\definecolor{NavyBlue}{RGB}{65, 105, 225}

\title{No Pose, No Problem: Surprisingly Simple 3D Gaussian Splats from Sparse Unposed Images}

\iclrfinalcopy

\author{Botao Ye$^{1}$ \qquad \qquad Sifei Liu$^{2}$ \qquad \qquad Haofei Xu$^{1}$ \qquad \qquad Xueting Li$^{2}$ \\[4pt]
\textbf{Marc Pollefeys}$^{1, 3}$ \qquad \qquad \textbf{Ming-Hsuan Yang}$^{4}$ \qquad \qquad \textbf{Songyou Peng}$^{1}$\thanks{Songyou Peng is currently at Google DeepMind, with this work mainly done at ETH Zurich.} \\[4pt]
$^1$ETH Zurich \quad  $^2$NVIDIA \quad $^3$Microsoft \quad $^4$UC Merced \\
\url{https://noposplat.github.io}
}

\iclrfinalcopy %
\begin{document}

\maketitle

\begin{center}
    \centering
    \captionsetup{type=figure}
    \includegraphics[width=.9\textwidth]{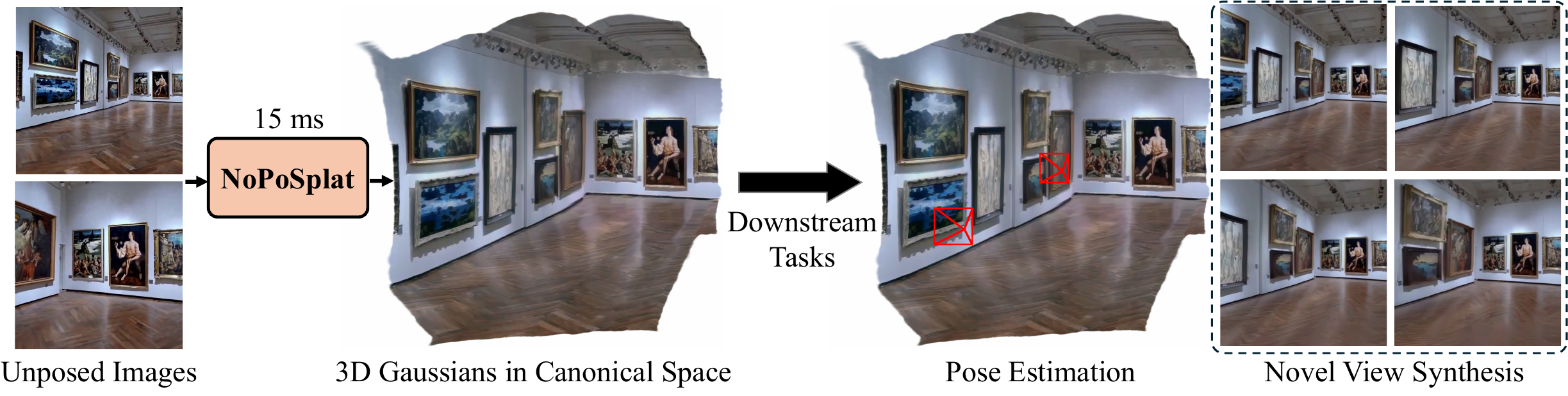}
    \captionof{figure}{\textbf{NoPoSplat}. Given sparse \emph{unposed} images, our method reconstructs 3D Gaussians of different views in a canonical space using a feed-forward network. The resulting 3D Gaussians can be utilized for accurate relative camera pose estimation and high-quality novel view synthesis. The input images for illustration are extracted from a Sora-generated video.}
\end{center}

\begin{abstract}
We introduce NoPoSplat, a feed-forward model capable of reconstructing 3D scenes parameterized by 3D Gaussians from \textit{unposed} sparse multi-view images. Our model, trained exclusively with photometric loss, achieves real-time 3D Gaussian reconstruction during inference. To eliminate the need for accurate pose input during reconstruction, we anchor one input view's local camera coordinates as the canonical space and train the network to predict Gaussian primitives for all views within this space. This approach obviates the need to transform Gaussian primitives from local coordinates into a global coordinate system, thus avoiding errors associated with per-frame Gaussians and pose estimation. To resolve scale ambiguity, we design and compare various intrinsic embedding methods, ultimately opting to convert camera intrinsics into a token embedding and concatenate it with image tokens as input to the model, enabling accurate scene scale prediction. We utilize the reconstructed 3D Gaussians for novel view synthesis and pose estimation tasks and propose a two-stage coarse-to-fine pipeline for accurate pose estimation.
Experimental results demonstrate that our pose-free approach can achieve superior novel view synthesis quality compared to pose-required methods, particularly in scenarios with limited input image overlap. 
For pose estimation, our method, trained without ground truth depth or explicit matching loss, significantly outperforms the state-of-the-art methods with substantial improvements. 
This work makes significant advances in pose-free generalizable 3D reconstruction and demonstrates its applicability to real-world scenarios.
Code and trained models are available on our \href{https://noposplat.github.io/}{project page}.
\end{abstract}

\section{Introduction}
We address the problem of reconstructing a 3D scene parameterized by 3D Gaussians from \textit{unposed} sparse-view images (as few as two) using a feed-forward network. 
While current SOTA generalizable 3D reconstruction methods~\citep{pixelsplat, mvsplat}, which aim to predict 3D radiance fields using feed-forward networks, can achieve photorealistic results without per-scene optimization, they require accurate camera poses of input views as input to the network.
These poses are typically obtained from dense videos using structure-from-motion (SfM) methods, such as COLMAP~\citep{colmap}.
This requirement is impractical for real-world applications, as these methods necessitate poses from dense videos even if only two frames are used for 3D reconstruction.
Furthermore, relying on off-the-shelf pose estimation methods increases inference time and can fail in textureless areas or images without sufficient overlap.

Recent methods~\citep{dbarf, flowcam, coponerf} aim to address this challenge by integrating pose estimation and 3D scene reconstruction into a single pipeline. 
However, the quality of novel view renderings from these methods lags behind SOTA approaches that rely on known camera poses~\citep{mvsplat}. The performance gap stems from their sequential process of alternating between pose estimation and scene reconstruction. Errors in pose estimation degrade the reconstruction, which in turn leads to further inaccuracies in pose estimation, creating a compounding effect.
In this work, we demonstrate that reconstructing the scene entirely without relying on camera poses is feasible, thereby eliminating the need for pose estimation. 
We achieve this by directly predicting scene Gaussians in a canonical space, inspired by the success of the recent 3D point cloud reconstruction methods~\citep{dust3r, mast3r}. 
However, unlike DUSt3R, we show that the generalizable reconstruction network can be trained with photometric loss only without ground truth depth information and thus can leverage more widely available video data~\citep{re10k, acid, dl3dv}.

\begin{figure*}[t!]
  \centering
  \includegraphics[width=\linewidth]{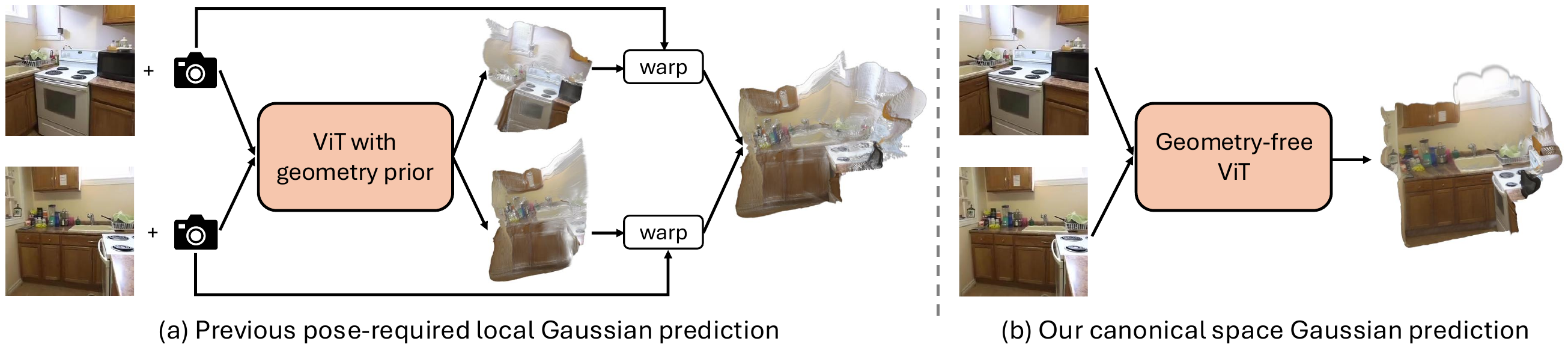}
  \caption{\textbf{Comparison with pose-required sparse-view 3D Gaussian splatting pipeline.} Previous methods first generate Gaussians in each local camera coordinate system and then transform them into a world coordinate system using camera poses. In contrast, NoPoSplat directly outputs 3D Gaussians of all views in a \textit{canonical} space, facilitating a more coherent fusion of multi-view geometric content (see Sec.~\ref{sec:unified_space} for details). Furthermore, our backbone does not incorporate geometry priors that rely on substantial image overlap (such as epipolar geometry in pixelSplat~\citep{pixelsplat} or cost volume in MVSplat~\citep{mvsplat}). Consequently, NoPoSplat demonstrates better geometric detail preservation when the overlap between input views is limited.}
  \label{fig:pipeline_comparison}
\end{figure*}

Specifically, as shown in Fig.~\ref{fig:pipeline_comparison} (b), we anchor the local camera coordinate of the first view as the canonical space and predict the Gaussian primitives~\citep{3dgs} for all input views within this space. Consequently, the output Gaussians will be aligned to this canonical space. This contrasts with previous methods~\citep{pixelsplat, mvsplat}, as illustrated in Fig.~\ref{fig:pipeline_comparison} (a), where Gaussian primitives are first predicted in each local coordinate system and then transformed to the world coordinate using the camera pose and fused together.
Compared to the transform-then-fuse pipeline, we require the network to learn the fusion of different views directly within the canonical space, thereby eliminating misalignments introduced by explicit transformations (see Fig.~\ref{fig:geometry}). 

Although the proposed pose-free pipeline is simple and promising, we observe significant scale misalignment in the rendered novel views compared to the ground truth target views (see Fig.~\ref{fig:ablation_vis}), \ie, the scene scale ambiguity issue. Upon analyzing the image projection process, we find that the camera's focal length is critical to resolving this scale ambiguity. This is because the model reconstructs the scene solely based on the image appearance, which is influenced by the focal length. Without incorporating this information, the model struggles to recover the scene at the correct scale.
To address this issue, we design and compare three different methods for embedding camera intrinsics and find that simply converting the intrinsic parameters into a feature token and concatenating it with the input image tokens enables the network to predict the scene of a more reasonable scale, yielding the best performance.

Once the 3D Gaussians are reconstructed in the canonical space, we leverage it for both novel view synthesis (NVS) and pose estimation. For pose estimation, we introduce a two-stage pipeline: first, we obtain an initial pose estimate by applying the PnP algorithm~\citep{hartley2003multiple} to the Gaussian centers. This rough estimate is then refined by rendering the scene at the estimated pose and optimizing the alignment with the input view using photometric loss.

Extensive experimental results demonstrate that our method performs impressively in both NVS and pose estimation tasks. For NVS, we show for the first time that, when trained on the same dataset under the same settings, a pose-free method can outperform pose-dependent methods, especially when the overlap between the two input images is small.
In terms of pose estimation, our method significantly outperforms prior SOTA across multiple benchmarks. 
Additionally, NoPoSplat generalizes well to out-of-distribution data. Since our method does not require camera poses for input images, it can be applied to user-provided images to reconstruct the underlying 3D scene and render novel views. To illustrate this, we apply our model to sparse image pairs captured with mobile phones, as well as to sparse frames extracted from videos generated by Sora~\citep{sora}.

The main contributions of this work are:
\begin{compactitem}
    \item We propose NoPoSplat, a feed-forward network that reconstructs 3D scenes parameterized by 3D Gaussians from unposed sparse-view inputs, and demonstrate that it can be trained using photometric loss alone.
    \item We investigate the scale ambiguity issue of the reconstructed Gaussians, and solve this problem by introducing a camera intrinsic token embedding.
    \item We design a two-stage pipeline that estimates accurate relative camera poses using the reconstructed Gaussian field.
    \item Experimental results show that our method achieves remarkable performance in both NVS and pose estimation tasks, and can be applied to in-the-wild data.
\end{compactitem}

\section{Related Work}
\boldparagraph{Generalizable 3D Reconstruction and View Synthesis}
NeRF~\citep{nerf} and 3D Gaussian Splatting (3DGS)~\citep{3dgs} have significantly advanced 3D reconstruction and novel view synthesis. However, these methods typically require dense posed images (\eg, hundreds) as input and minutes to hours of per-scene optimization, even with improved techniques~\citep{tensorf, kplanes, instant-ngp}. 
This limits their practical applications.
To address these limitations, recent approaches focus on generalizable 3D reconstruction and novel view synthesis from sparse inputs~\citep{pixelnerf, ibrnet, murf, geonerf, du2023learning, pixelsplat, mvsplat}.
They typically incorporate task-specific backbones that leverage geometric information to enhance scene reconstruction. For instance, MVSNeRF~\citep{mvsnerf} and MuRF~\citep{murf} build cost volumes to aggregate multi-view information, while pixelSplat~\citep{pixelsplat} employs epipolar geometry for improved depth estimation.
However, these geometric operations often require camera pose input and sufficient camera pose overlap among input views.
In contrast, our network is based solely on a vision transformer~\citep{vit} without any geometric priors, making it pose-free and more effective in handling scenes with large camera baselines.

\boldparagraph{Pose-Free 3D Scene Reconstruction}
Classical NeRF or 3DGS-based methods require accurate camera poses of input images, typically obtained through Structure from Motion (SfM) methods like COLMAP~\citep{colmap}, which complicates the overall process. 
Recent works~\citep{nerfmm, barf, garf}  jointly optimize camera poses and neural scene representations, but they still require rough pose initialization or are limited to small motions. Others~\citep{nope-nerf, colmap-free3DGS} adopt incremental approaches from~\citep{niceslam, nicerslam, monogs}, but they only allow image/video sequences as input. 

Moreover, for generalizable sparse-view methods, requiring camera poses during inference presents significant challenges, as these poses are often unavailable in real-world applications during testing. Although two-view pose estimation methods~\citep{dust3r, roma} can be used, they are prone to failure in textureless regions or when images lack sufficient overlap.
Some recent pose-free novel view synthesis methods~\citep{dbarf, flowcam, coponerf} attempt to address this but typically break the task into two stages: first estimate camera poses, then construct the scene representation. This two-stage process still lags behind pose-required methods because the initial pose estimation introduces noise, degrading reconstruction quality.
In contrast, our method completely eliminates camera poses by directly predicting 3D Gaussians in a canonical space, avoiding potential noise in pose estimation and achieving better scene reconstruction.

One concurrent work, Splatt3R~\citep{splatt3r}\footnote{Arxiv submitted on August 25, 2024}, also predicts Gaussians in a global coordinate system but relies on the frozen MASt3R~\citep{mast3r} model for Gaussian centers.
This is unsuitable for novel view synthesis, as MASt3R struggles to merge scene content from different views smoothly. 
Moreover, in contrast to us, Splatt3R requires ground truth depth during training, so it cannot utilize large-scale video data without depths or ground truth metric camera poses.

\section{Method}

\subsection{Problem Formulation}
\label{sec:problem_formulation}
Our method takes as input sparse \textit{unposed} multi-view images and corresponding camera intrinsic parameters $\left\{\boldsymbol{I}^v, \boldsymbol{k}^v\right\}_{v=1}^V$, where $V$ is the number of input views, and learn a feed-forward network $f_{\boldsymbol{\theta}}$ with learnable parameters $\theta$.
The network maps the input unposed images to 3D Gaussians in a \textit{canonical} 3D space, representing the underlying scene geometry and appearance. Formally, we aim to learn the following mapping:
\begin{equation}
f_{\boldsymbol{\theta}}:\left\{\left(\boldsymbol{I}^v, \boldsymbol{k}^v\right)\right\}_{v=1}^V \mapsto\left\{ \cup \left(\boldsymbol{\mu}_j^v, \alpha_j^v, \boldsymbol{r}_j^v, \boldsymbol{s}_j^v, \boldsymbol{c}_j^v\right)\right\}_{j=1,\dots,H \times W}^{v=1,\dots,V},
\label{eq:mapping}
\end{equation}
where the right side represents the Gaussian parameters~\citep{3dgs}. Specifically, we have the center position $\boldsymbol{\mu} \in \mathbb{R}^3$ , opacity $\alpha \in \mathbb{R}$,
rotation factor in quaternion $\boldsymbol{r} \in \mathbb{R}^4$,
scale $\boldsymbol{s} \in \mathbb{R}^3$, and  spherical harmonics (SH) $\boldsymbol{c} \in \mathbb{R}^k$ with $k$ degrees of freedom.
Note that, in line with common practice in pose-free scene reconstruction methods~\citep{coponerf, colmap-free3DGS, flowcam, dbarf}, we assume having camera intrinsic parameters $\boldsymbol{k}$ as input, as they are generally available from modern devices~\citep{arnold2022map}.

By training on large-scale datasets, our method can generalize to novel scenes without any optimization.
The resulting 3D Gaussians in the canonical space directly enable two tasks: a) novel view synthesis given the target camera transformation relative to the first input view, and b) relative pose estimation among different input views.
Next, we will introduce our overall pipeline.

\begin{figure*}[t]
    \centering
    \includegraphics[width=0.95\linewidth]{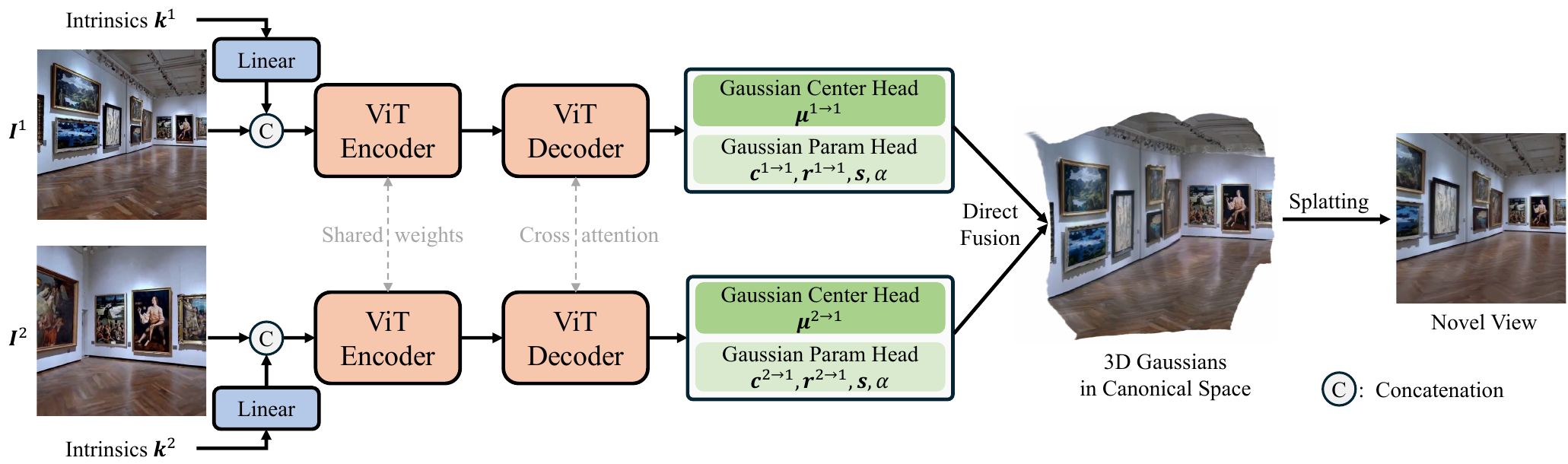}
    \vspace{-0.5em}\caption{\textbf{Overview of NoPoSplat.} We directly predict Gaussians in a canonical space from a feed-forward network to represent the underlying 3D scene from the unposed sparse images. For simplicity, we use a two-view setup as an example, and the RGB shortcut is omitted from the figure. 
    }
    \label{fig:pipeline}
\end{figure*}
\subsection{Pipeline}
\label{sec:network}
Our method, illustrated in Fig.~\ref{fig:pipeline}, comprises three main components: an encoder, a decoder, and Gaussian parameter prediction heads. Both encoder and decoder utilize pure Vision Transformer (ViT) structures, without injecting any geometric priors (e.g. epipolar constraints employed in pixelSplat~\citep{pixelsplat}, or cost volumes in MVSplat~\citep{mvsplat}). Interestingly, we demonstrate in~Sec.~\ref{sec:exp} that such a simple ViT network shows competitive or superior performance over those dedicated backbones incorporating these geometric priors, especially in scenarios with limited content overlap between input views. This advantage stems from the fact that such geometric priors typically necessitate substantial overlap between input cameras to be effective.

\boldparagraph{ViT Encoder and Decoder}
The RGB images are patchified and flattened into sequences of image tokens, and then concatenated with an intrinsic token (details in Sec.~\ref{subsec:intrinsic}). The concatenated tokens from each view are then fed into a ViT~\citep{vit} encoder separately. The encoder shares the same weights for different views.
Next, the output features from the encoder are fed into a ViT decoder module, where features from each view interact with those from all other views through cross-attention layers in each attention block, facilitating multi-view information integration.

\boldparagraph{Gaussian Parameter Prediction Heads} 
To predict the Gaussian parameters, we employ two prediction heads based on the DPT architecture~\citep{dpt}. The first head focuses on predicting the Gaussian center positions and utilizes features extracted exclusively from the transformer decoder.
The second head predicts the remaining Gaussian parameters and, in addition to the ViT decoder features, also takes the RGB image as input.
Such RGB image shortcut ensures the direct flow of texture information, which is crucial for capturing fine texture details in 3D reconstruction. 
This approach compensates for the high-level features output by the ViT decoder, downsampled by a factor of 16, which are predominantly semantic and lack detailed structural information.
With these prediction heads in place, we now analyze how our method differs from previous approaches in terms of the output Gaussian space and the advantages this brings.

\subsection{Analysis of the Output Gaussian Space}
\label{sec:unified_space}
While our method shares a similar spirit with previous works~\citep{pixelsplat, gps-gaussian, splatter-image} in predicting pixelwise Gaussians for input images, we differ in the output Gaussian space.  In this section, we first discuss the local-to-global Gaussian space in prior methods and its inherent limitations, and introduce our canonical Gaussian space.
\boldparagraph{Baseline: Local-to-Global Gaussian Space}
Previous methods first predict the corresponding depth of each pixel, then lift the predicted Gaussian parameters to a Gaussian primitive in the local coordinate system of each individual frame using the predicted depth and the camera intrinsics. 
These local Gaussians are then transformed into a world coordinate system using the given camera poses $[\boldsymbol{R}^v \mid \boldsymbol{t}^v]$ for each input view. Finally, all transformed Gaussians are fused to represent the underlying scene.

However, this strategy has two main issues:
a) Transforming Gaussians from local to world coordinates requires accurate camera poses, which are difficult to obtain in real-world scenarios with sparse input views.
b) The transform-then-fuse method struggles to combine 3D Gaussians from different views into a cohesive global representation, especially when the overlap among input views is small, or when generalizing to out-of-distribution data (as shown in Fig.~\ref{fig:geometry} and Fig.~\ref{fig:zero_shot}).

\boldparagraph{Proposed: Canonical Gaussian Space}
In contrast, we directly output Gaussians of different views in a canonical coordinate system.
Specifically, we anchor the first input view as the global reference coordinate system. 
Therefore, the camera pose for the first input view is $[\boldsymbol{U}\mid\boldsymbol{0}]$, where $\boldsymbol{U}$ is the unit/identity matrix for rotation, and $\boldsymbol{0}$ is the zero translation vector.
The network outputs Gaussians under this canonical space for all input views. Formally, for each input view, we predict the set $\left\{ \boldsymbol{\mu}^{v \rightarrow 1}_{j}, \boldsymbol{r}^{v \rightarrow 1}_{j}, \boldsymbol{c}^{v \rightarrow 1}_{j}, \alpha_{j}, \boldsymbol{s}_j \right\}$, where the superscript $v \rightarrow 1$ denotes that the Gaussian parameters corresponding to pixel $\boldsymbol{p}_j$ in view $v$, are under the local camera coordinate system of view $1$.

Predicting directly under the canonical space offers several benefits.
First, the network learns to fuse different views directly within the canonical space, eliminating the need for camera poses.
Second, bypassing the transform-then-fuse step results in a cohesive global representation, which further unlocks the application of pose estimation for input unposed views.

\subsection{Camera Intrinsics Embedding}
\label{subsec:intrinsic}
As discussed in Eq.~\ref{eq:mapping}, our network inputs also include the camera intrinsics $\boldsymbol{k}$ of each input view.
This is required to resolve the inherent scale misalignment and provide essential geometric information that improves 3D reconstruction quality (\cf. ``No Intrinsics'' in Fig.~\ref{fig:ablation_vis} and Tab.~\ref{tab:ablation}).
We introduce three different encoding strategies for injecting camera intrinsics into our model.
\boldparagraph{Global Intrinsic Embedding - Addition}
A straightforward strategy is to feed camera intrinsics $\boldsymbol{k}=[f_x, f_y, c_x, c_y]$ into a linear layer to obtain a global feature. 
This feature is broadcast and added to the RGB image features after the patch embedding layer of the ViT.

\boldparagraph{Global Intrinsic Embedding - Concat}
After obtaining the global feature, we instead treat it as an additional intrinsic token, and concatenate it with all image tokens.

\boldparagraph{Dense Intrinsic Embedding}
For each pixel $\boldsymbol{p_j}$ in the input view, we can obtain the camera ray direction as $\boldsymbol{K}^{-1}\boldsymbol{p_j}$, where $\boldsymbol{K}$ is the matrix form of $\boldsymbol{k}$. These per-pixel camera rays are then converted using spherical harmonics to higher-dimension features and concatenated with the RGB image as the network input.
Note that the pixel-wise embedding can be viewed as a simplification of the widely-used Plücker ray embedding~\citep{dmv3d, lgm}, as it does not require camera extrinsic information.

By default, we utilize the intrinsic token option since it is not only an elegant way to inject camera intrinsic information into the network, but also yields the best performance. (see Tab.~\ref{tab:ablation}).

\subsection{Training and Inference}
\label{subsec:train_test}

\boldparagraph{Training} 
The predicted 3D Gaussians are used for rendering images at novel viewpoint.
Our network is end-to-end trained using ground truth target RGB images as supervision.
Following~\cite{mvsplat}, we also use a linear combination of MSE and LPIPS~\citep{lpips} loss with weights of 1 and 0.05, respectively.

\boldparagraph{Relative Pose Estimation}
As mentioned in Sec.~\ref{sec:unified_space}, since our 3D Gaussians are in the canonical space, they can be directly used for relative pose estimation. 
To facilitate efficient pose estimation, we propose a two-step approach.
First, we estimate the initial related camera pose of the input two views using the PnP algorithm~\citep{hartley2003multiple} with RANSAC~\citep{ransac}, given the Gaussian centers of the output Gaussians in world coordinates. This step is highly efficient and done in milliseconds.
Next, while keeping Gaussian parameters frozen, we refine the initial pose from the first step by optimizing the same photometric losses used for model training, along with the structural part of the SSIM loss~\citep{ssim}. 
During the optimization, we calculate the camera Jacobian to reduce the computational overhead associated with automatic differentiation and decrease optimization time as in~\cite{monogs}.

\boldparagraph{Evaluation-Time Pose Alignment} 
Given unposed image pairs, our method learns to reconstruct a plausible 3D scene that aligns with the given inputs.
However, 3D scene reconstruction with just two input views is inherently ambiguous as many different scenes can produce the same two images.
As a result, though the scene generated by our method successfully explains the input views, it might not be exactly the same as the ground truth scene in the validation dataset.
Thus, to fairly compare with other baselines, especially ones that utilize ground truth camera poses\citep{du2023learning, pixelsplat}, we follow previous pose-free works~\citep{nerfmm,instantsplat} and optimize the camera pose for the target view.
Specifically, for each evaluation sample, we first reconstruct 3D Gaussians using the proposed method. We then freeze the Gaussians and optimize the target camera pose such that the rendered image from the target view closely matches the ground truth image. 
It is important to note that this optimization is solely for evaluation purposes and is not required when applying our method in real-world scenarios (\eg, Fig.~\ref{fig:wild_test}).

\section{Experiments}
\label{sec:exp}

\textbf{Datasets.}
To evaluate novel view synthesis, we follow the setting in~\citep{pixelsplat, mvsplat} and train and evaluate our method on RealEstate10k (RE10K)~\citep{re10k} and ACID~\citep{acid} datasets separately.
RE10K primarily contains indoor real estate videos, while ACID features nature scenes captured by aerial drones. Both include camera poses calculated using COLMAP~\citep{colmap}. We adhere to the official train-test split as in previous works~\citep{pixelsplat, coponerf}.
To further scale up our model (denoted as \textbf{Ours*}), we also combine RE10K with DL3DV~\citep{dl3dv}, which is an outdoor dataset containing 10K videos, which includes a wider variety of camera motion patterns than RE10K.

To assess the method's capability in handling input images with varying camera overlaps, we generate input pairs for evaluation that are categorized based on the ratio of image overlap:
small (0.05\% - 0.3\%),  medium (0.3\% - 0.55\%), and large (0.55\% - 0.8\%), determined using SOTA dense feature matching method, RoMA~\citep{roma}. More details are provided in the appendix.

Furthermore, for zero-shot generalization, we also test on
DTU~\citep{dtu} (object-centric scenes), ScanNet~\citep{scannet} and ScanNet++~\citep{scannetpp} (indoor scenes with different camera motion and scene types from the RE10K).
We also demonstrate our approach on in-the-wild mobile phone capture, and images generated by a text-to-video model~\citep{sora}.

\textbf{Evaluation Metrics.}
We evaluate novel view synthesis with the commonly used metrics: PSNR, SSIM, and LPIPS~\citep{lpips}. 
For pose estimation, we report the area under the cumulative pose error curve (AUC) with thresholds of 5$^\circ$, 10$^\circ$, 20$^\circ$~\citep{superglue, roma}. 

\textbf{Baselines.}
We compare against SOTA sparse-view generalizable methods on novel view synthesis:
1) \emph{Pose-required}: pixelNeRF~\citep{pixelnerf}, AttnRend~\citep{du2023learning}, pixelSplat~\citep{pixelsplat}, and MVSplat~\citep{mvsplat};
2) \emph{Pose-free}: DUSt3R~\citep{dust3r}, MASt3R~\citep{mast3r}, Splatt3R~\citep{splatt3r}, CoPoNeRF~\citep{coponerf}, and RoMa~\citep{roma}.
For relative pose estimation, we also compare against methods in 2).

\begin{table}[!t]
    \centering
    \caption{\textbf{Novel view synthesis performance comparison on the RealEstate10k~\citep{re10k} dataset}. 
    Our method largely outperforms previous pose-free methods on all overlap settings, and even outperforms SOTA pose-required methods, especially when the overlap is small.}
    \begin{adjustbox}{max width=\textwidth}
    \setlength{\tabcolsep}{0.08cm}
    \renewcommand{\arraystretch}{0.5}
    \begin{tabular}{llccc ccc ccc ccc}
        \toprule
        & & \multicolumn{3}{c}{Small} & \multicolumn{3}{c}{Medium} & \multicolumn{3}{c}{Large} & \multicolumn{3}{c}{Average} \\
        \cmidrule(lr){3-5} \cmidrule(lr){6-8} \cmidrule(lr){9-11} \cmidrule(lr){12-14}
        & Method & PSNR$\uparrow$ & SSIM$\uparrow$ & LPIPS$\downarrow$ & PSNR$\uparrow$ & SSIM$\uparrow$ & LPIPS$\downarrow$ & PSNR$\uparrow$ & SSIM$\uparrow$ & LPIPS$\downarrow$ & PSNR$\uparrow$ & SSIM$\uparrow$ & LPIPS$\downarrow$ \\
        \midrule
        \multirow{4}{*}{\shortstack[l]{\emph{Pose-} \\ \emph{Required}}} 
        & pixelNeRF & 18.417 & 0.601 & 0.526 & 19.930 & 0.632 & 0.480 & 20.869 & 0.639 & 0.458 & 19.824 & 0.626 & 0.485 \\
        & AttnRend & 19.151 & 0.663 & 0.368 & 22.532 & 0.763 & 0.269 & 25.897 & 0.845 & 0.186 & 22.664 & 0.762 & 0.269 \\
        & pixelSplat & 20.263 & 0.717 & 0.266 & 23.711 & 0.809 & 0.181 & 27.151 & 0.879 & 0.122 & 23.848 & 0.806 & 0.185 \\
        & MVSplat & 20.353 & 0.724 & 0.250 & 23.778 & 0.812 & 0.173 & 27.408 & \best{0.884} & \best{0.116} & 23.977 & 0.811 & 0.176 \\
        \midrule
        \multirow{5}{*}{\shortstack[l]{\emph{Pose-} \\ \emph{Free}}} 
        & DUSt3R & 14.101 & 0.432 & 0.468 & 15.419 & 0.451 & 0.432 & 16.427 & 0.453 & 0.402 & 15.382 & 0.447 & 0.432 \\
        & MASt3R & 13.534 & 0.407 & 0.494 & 14.945 & 0.436 & 0.451 & 16.028 & 0.444 & 0.418 & 14.907 & 0.431 & 0.452 \\
        & Splatt3R & 14.352 & 0.475 & 0.472 & 15.529 & 0.502 & 0.425 & 15.817 & 0.483 & 0.421 & 15.318 & 0.490 & 0.436 \\
        & CoPoNeRF & 17.393 & 0.585 & 0.462 & 18.813 & 0.616 & 0.392 & 20.464 & 0.652 & 0.318 & 18.938 & 0.619 & 0.388 \\
        & \best{Ours} & \best{22.514} & \best{0.784} & \best{0.210} & \best{24.899} & \best{0.839} & \best{0.160} & \best{27.411} & 0.883 & 0.119 & \best{25.033} & \best{0.838} & \best{0.160} \\
        \bottomrule
    \end{tabular}
    \end{adjustbox}
\label{tab:re10k}
\end{table}

\begin{table}[t]
    \centering
    \caption{\textbf{Novel view synthesis performance comparison on the ACID~\citep{acid} dataset}.}
    \begin{adjustbox}{max width=\textwidth}
    \setlength{\tabcolsep}{0.08cm}
    \renewcommand{\arraystretch}{0.5}
    \begin{tabular}{llccc ccc ccc ccc}
        \toprule
        & & \multicolumn{3}{c}{Small} & \multicolumn{3}{c}{Medium} & \multicolumn{3}{c}{Large} & \multicolumn{3}{c}{Average} \\
        \cmidrule(lr){3-5} \cmidrule(lr){6-8} \cmidrule(lr){9-11} \cmidrule(lr){12-14}
        & Method & PSNR$\uparrow$ & SSIM$\uparrow$ & LPIPS$\downarrow$ & PSNR$\uparrow$ & SSIM$\uparrow$ & LPIPS$\downarrow$ & PSNR$\uparrow$ & SSIM$\uparrow$ & LPIPS$\downarrow$ & PSNR$\uparrow$ & SSIM$\uparrow$ & LPIPS$\downarrow$ \\
        \midrule
        \multirow{4}{*}{\shortstack[l]{\emph{Pose-} \\ \emph{Required}}} 
        & pixelNeRF & 19.376 & 0.535 & 0.564 & 20.339 & 0.561 & 0.537 & 20.826 & 0.576 & 0.509 & 20.323 & 0.561 & 0.533 \\
        & AttnRend & 20.942 & 0.616 & 0.398 & 24.004 & 0.720 & 0.301 & 27.117 & 0.808 & 0.207 & 24.475 & 0.730 & 0.287 \\
        & pixelSplat & 22.053 & 0.654 & 0.285 & 25.460 & 0.776 & 0.198 & \best{28.426} & \best{0.853} & 0.140 & 25.819 & 0.779 & 0.195 \\
        & MVSplat & 21.392 & 0.639 & 0.290 & 25.103 & 0.770 & 0.199 & 28.388 & 0.852 & \best{0.139} & 25.512 & 0.773 & 0.196 \\
        \midrule
        \multirow{4}{*}{\shortstack[l]{\emph{Pose-} \\ \emph{Free}}} 
        & DUSt3R & 14.494 & 0.372 & 0.502 & 16.256 & 0.411 & 0.453 & 17.324 & 0.431 & 0.408 & 16.286 & 0.411 & 0.447 \\
        & MASt3R & 14.242 & 0.366 & 0.522 & 16.169 & 0.411 & 0.463 & 17.270 & 0.430 & 0.423 & 16.179 & 0.409 & 0.461 \\
        & CoPoNeRF & 18.651 & 0.551 & 0.485 & 20.654 & 0.595 & 0.418 & 22.654 & 0.652 & 0.343 & 20.950 & 0.606 & 0.406 \\
        & \textbf{Ours} & \best{23.087} & \best{0.685} & \best{0.258} & \best{25.624} & \best{0.777} & \best{0.193} & 28.043 & 0.841 & 0.144 & \best{25.961} & \best{0.781} & \best{0.189} \\
        \bottomrule
    \end{tabular}
    \end{adjustbox}
\label{tab:acid}
\end{table}

\textbf{Implementation details.}
We use PyTorch, and the encoder is a vanilla ViT-large model with a patch size of 16, and the decoder is ViT-base.
We initialize the encoder/decoder and Gaussian center head with the weights from MASt3R, while the remaining layers are initialized randomly. 
Note that, as shown in the appendix, our method can also be trained %
with only RGB supervision--without pre-trained weight from MASt3R--and still achieve similar performance. 
We train models at two different resolutions, $256\times256$ and $512\times512$. For a fair comparison with baseline models, we report all quantitative results and baseline comparisons under $256\times256$. However, qualitative results for the $512\times512$ model are presented in the supplementary video and Fig.~\ref{fig:wild_test}. Additional details on model weight initialization and training resolution can be found in the appendix.

\subsection{Experimental Results and Analysis}
\begin{figure*}[t]
    \centering
    \fontsize{9}{11}\selectfont
    \includeinkscape{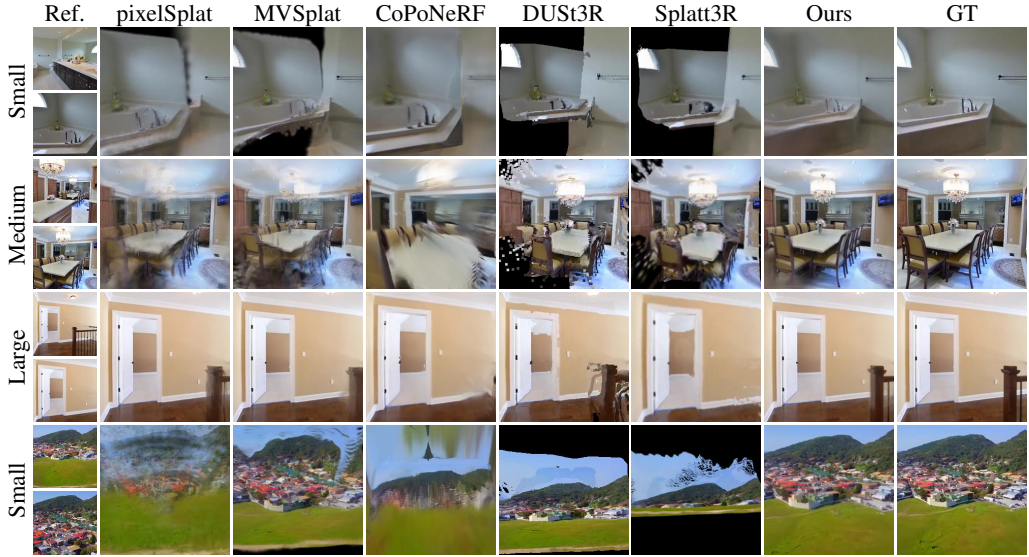}
    \caption{\textbf{Qualitative comparison on RE10K (top three rows) and ACID (bottom row).}
    Compared to baselines, we obtain: 1) more coherent fusion from input views, 2) superior reconstruction from limited image overlap, 3) enhanced geometry reconstruction in non-overlapping regions.
    }
    \label{fig:sota_comparison}
\end{figure*}

\textbf{Novel View Synthesis.}
As demonstrated in Tab.~\ref{tab:re10k}, Tab.~\ref{tab:acid}, and Fig.~\ref{fig:sota_comparison}, NoPoSplat significantly outperforms all SOTA pose-free approaches. 
Note that DUSt3R (and MASt3R) struggle to fuse input views coherently due to their reliance on per-pixel depth loss, a limitation Splatt3R also inherits from its frozen MASt3R module.
On the other hand, we achieve competitive performance over SOTA pose-required methods~\citep{pixelsplat, mvsplat}, and even outperform them when the overlap between input images is small, as shown in Fig.\ref{fig:sota_comparison} first row (the left side of Fig.~\ref{fig:sota_comparison} displays the overlap ratio). 
This clearly shows the advantages of our model's 3D Gaussians prediction in the canonical space over baseline methods' transform-then-fuse strategy, as discussed in Sec.~\ref{sec:unified_space}.

\textbf{Relative Pose Estimation.}
\begin{table}[t]
\centering
\caption{\textbf{Pose estimation performance in AUC with various thresholds on RE10k, ACID, and ScanNet-1500 \citep{scannet, superglue}.} Our method achieves the best results across all datasets. Notably, our model is not trained on ACID or ScanNet. Furthermore, our method does not require an explicit matching loss during training, meaning no ground truth depth is necessary.
}
\label{tab:pose_compare_all}
\begin{adjustbox}{max width=0.8\textwidth}
\begin{tabular}{lccccccccc}
\toprule
& \multicolumn{3}{c}{RE10k} & \multicolumn{3}{c}{ACID} & \multicolumn{3}{c}{ScanNet-1500} \\
\cmidrule(lr){2-4} \cmidrule(lr){5-7} \cmidrule(lr){8-10}
Method & 5$^\circ\uparrow$ & 10$^\circ\uparrow$ & 20$^\circ\uparrow$ & 5$^\circ\uparrow$ & 10$^\circ\uparrow$ & 20$^\circ\uparrow$ & 5$^\circ\uparrow$ & 10$^\circ\uparrow$ & 20$^\circ\uparrow$ \\
\midrule
CoPoNeRF & 0.161 & 0.362 & 0.575 & 0.078 & 0.216 & 0.398 & - & - & - \\
DUSt3R & 0.301 & 0.495 & 0.657 & 0.166 & 0.304 & 0.437 & 0.221 & 0.437 & 0.636 \\
MASt3R & 0.372 & 0.561 & 0.709 & 0.234 & 0.396 & 0.541 & 0.159 & 0.359 & 0.573 \\
RoMa & 0.546 & 0.698 & 0.797 & 0.463 & 0.588 & 0.689 & \second{0.270} & \second{0.492} & \second{0.673} \\
\textbf{Ours} (RE10k) & \second{0.672} & \second{0.792} & \second{0.869} & \second{0.454} & \second{0.591} & \second{0.709} & 0.264 & 0.473 & 0.655 \\
\textbf{Ours*} (RE10k+DL3DV) & \best{0.691} & \best{0.806} & \best{0.877} & \best{0.486} & \best{0.617} & \best{0.728} & \best{0.318} & \best{0.538} & \best{0.717} \\
\bottomrule
\end{tabular}%
\end{adjustbox}
\end{table}

The proposed method can be applied to pose estimation between input views on three diverse datasets.
Our method is trained either on RE10K (denoted as Ours) or a combination of RE10K and DL3DV (denoted as Ours*). 
Tab.~\ref{tab:pose_compare_all} shows that the performance consistently improves when scaling up training with DL3DV involved. 
This can be attributed to the greater variety of camera motions in DL3DV over in RE10K.
It is worth noting that our method even shows superior zero-shot performance on ACID and ScanNet-1500, even better than the SOTA method RoMa that is trained on ScanNet. This indicates not only the efficacy of our pose estimation approach, but also the quality of our output 3D geometry. The next part verifies this point.

\textbf{Geometry Reconstruction.}
\begin{figure*}[t]
    \centering
    \includegraphics[width=\linewidth]{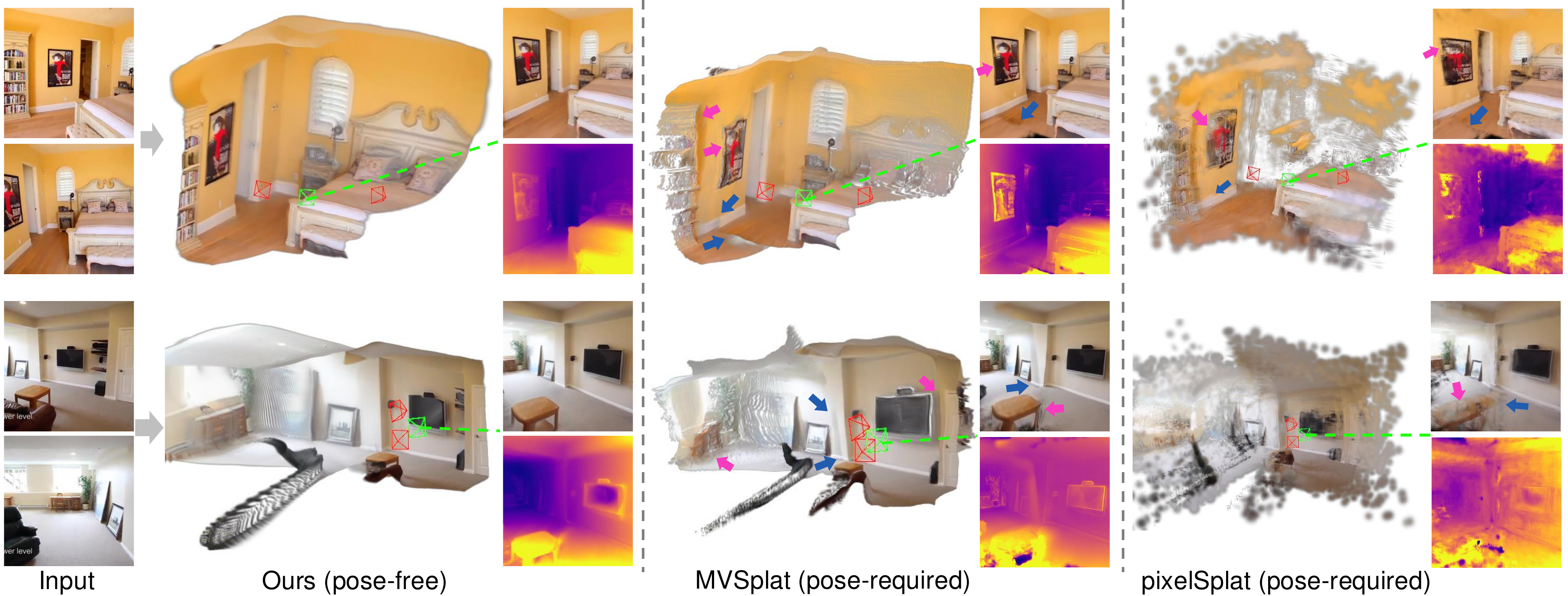}
    \caption{\textbf{Comparisons of 3D Gaussian and rendered results.} 
    The {\color{red}red} and {\color{green}green} indicate input and target camera views, and the rendered image and depths are shown on the right side. The {\color{magenta}magenta} and {\color{NavyBlue}blue} arrows correspond to the distorted or misalignment regions in baseline 3DGS.
    The results show that even without camera poses as input, our method produces higher-quality 3D Gaussians resulting in better color/depth rendering over baselines.
    }
    \label{fig:geometry}
\end{figure*}
Our method also outputs noticeably better 3D Gaussians and depths over SOTA pose-required methods, as shown in Fig.~\ref{fig:geometry}.
Looking closely, MVSplat not only suffers from the misalignment in the intersection regions of two input images (indicated by blue arrows), but also distortions or incorrect geometry in regions without sufficient overlap (indicated by magenta arrows).
These issues are largely due to the noises introduced in their transform-then-fuse pipeline.
Our method directly predicts Gaussians in the canonical space, which faithfully solves this problem. 

\begin{figure*}[t!]
    \centering
    \vspace{0.3em}
    \begin{subfigure}[t]{0.5\textwidth}
    \fontsize{9}{11}\selectfont
        \centering
        \includeinkscape[width=1.0\linewidth]{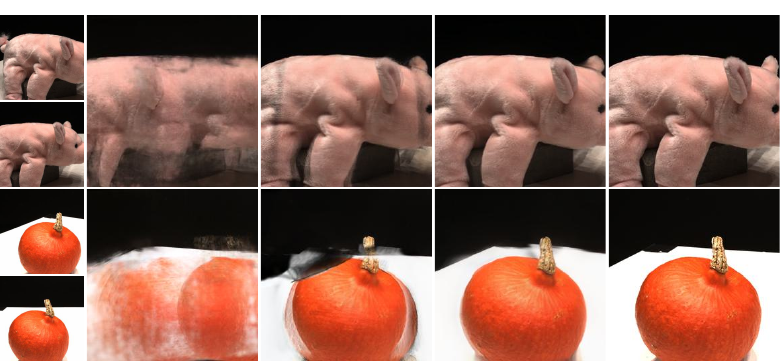}
        \caption{Cross-Dataset Generalize: RE10K $\rightarrow$ DTU}
        \label{fig:zero_dtu}
    \end{subfigure}%
    ~ 
    \begin{subfigure}[t]{0.5\textwidth}
    \fontsize{9}{11}\selectfont
        \centering
        \includeinkscape[width=1.0\linewidth]{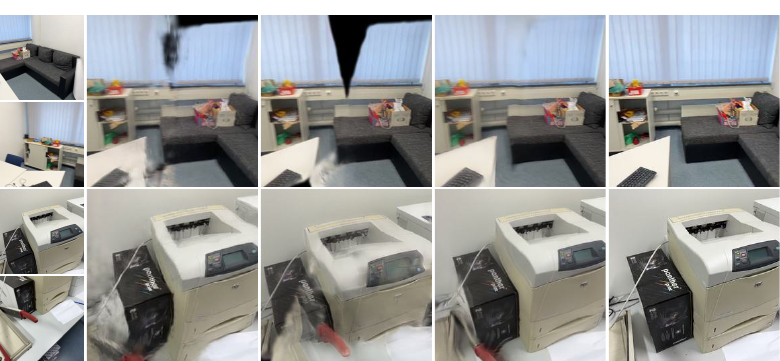}
        \caption{Cross-Dataset Generalize: RE10K $\rightarrow$ ScanNet++}
        \label{fig:zero_scannetpp}
    \end{subfigure}
    \caption{\textbf{Cross-dataset generalization.} Our model can better zero-shot transfer to out-of-distribution data than SOTA pose-required methods. }
    \label{fig:zero_shot}
\end{figure*}
\begin{table}[t!]
    \centering
    \caption{\textbf{Out-of-distribution performance comparison.} Our method shows superior performance when zero-shot evaluation on DTU and ScanNet++ using the model solely trained on RE10k.}
    \begin{adjustbox}{max width=0.7\textwidth}
    \begin{tabular}{lccccccc}
        \toprule
        \multirow{2}{*}{Training Data} & \multirow{2}{*}{Method} & \multicolumn{3}{c}{DTU} & \multicolumn{3}{c}{ScanNet++} \\
        \cmidrule(lr){3-5} \cmidrule(lr){6-8}
         &  & PSNR $\uparrow$ & SSIM $\uparrow$ & LPIPS $\downarrow$ & PSNR $\uparrow$ & SSIM $\uparrow$ & LPIPS $\downarrow$ \\
        \midrule

        \multirow{1}{*}{ScanNet++}
        & Splatt3R & 11.628 & 0.325 & 0.499 & 13.270 & 0.507 & 0.445 \\

        \midrule

        \multirow{3}{*}{RealEstate10K} 
         & pixelSplat & 11.551 & 0.321 & 0.633 & 18.434 & 0.719 & 0.277 \\
         & MVSplat    & 13.929 & 0.474 & 0.385 & 17.125 & 0.686 & 0.297 \\
         & Ours       & \best{17.899} & \best{0.629} & \best{0.279} & \best{22.136} & \best{0.798} & \best{0.232} \\
        \bottomrule
    \end{tabular}
    \end{adjustbox}
    \label{tab:zero_shot}
\end{table}

\textbf{Cross-Dataset Generalization.}
We also evaluate the zero-shot performance of the model, where we train exclusively on RealEstate10k and directly apply it to ScanNet++ \citep{scannetpp} and DTU \citep{dtu} datasets. The results in Fig.~\ref{fig:zero_shot} and Tab.~\ref{tab:zero_shot} indicate that NoPoSplat demonstrates superior performance on out-of-distribution data compared to SOTA pose-required methods. This advantage arises primarily from our minimal geometric priors in the network structure, allowing it to adapt effectively to various types of scenes. Notably, our method outperforms Splatt3R even on ScanNet++, where Splatt3R was trained. 

\begin{wrapfigure}[4]{r}{0.25\textwidth} 
    \centering
    \vspace{-1.5em} %
    \includegraphics[width=0.25\textwidth]{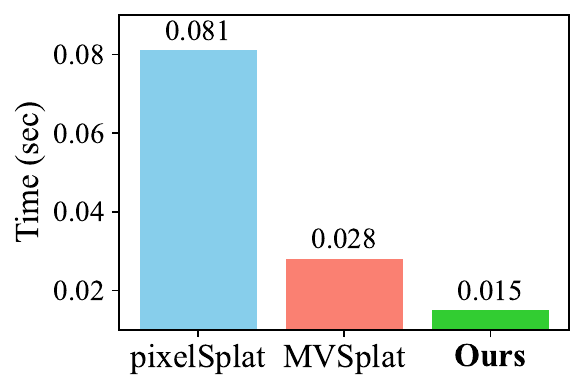}
    \label{fig:prediction_time}
\end{wrapfigure}

\textbf{Model Efficiency.}
As shown on the right, our method can predict 3D Gaussians from two $256\times256$ input images in 0.015 seconds (66 fps), which is around $5\times$ and $2\times$ faster than pixelSplat and MVSplat, on the same RTX 4090 GPU.
This further shows the benefits of using a standard ViT without incorporating additional geometric operations.
\begin{figure*}[t]
    \centering
    \includegraphics[width=\linewidth]{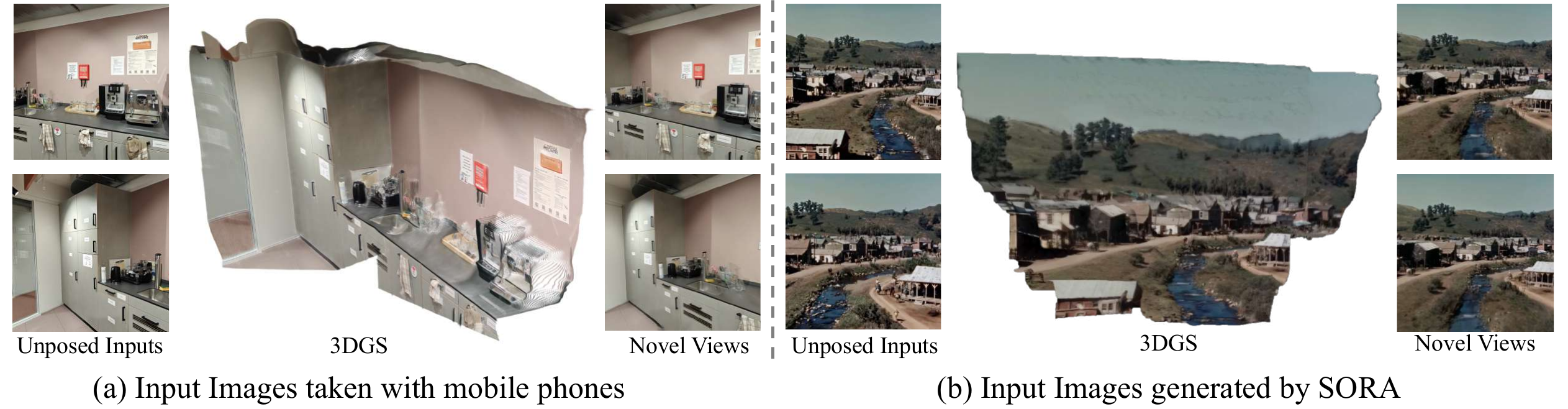}
    \vspace{-1.5em}
    \caption{\textbf{In-the-wild Data.} We present the results of applying our method to in-the-wild data, including real-world photos taken with mobile phones and multi-view images extracted from videos generated by the Sora text-to-video model. See the video supplementary for more results.}
    \label{fig:wild_test}
\end{figure*}

\boldparagraph{Apply to In-the-Wild Unposed Images}
One of the most important advantages of our method is that it can directly generalize to in-the-wild unposed images. 
we test on two types of data: images casually taken with mobile phones, and image frames extracted from videos generated by Sora~\citep{sora}.
Results in Fig.~\ref{fig:wild_test} show that our method can be potentially applied for text/image to 3D scene generations, \eg, one can first generate sparse scene-level multi-view images using text/image to multi-image/video models~\citep{cat3d, sora}, then feed those extracted unposed images to our model and obtain 3D models.

\subsection{Ablation Studies}

\textbf{Ablation on Output Gaussian Space.}
To demonstrate the effectiveness of our canonical Gaussian prediction, we compare it with the transform-then-fuse pipeline commonly used by pose-required methods~\citep{pixelsplat,mvsplat}.
Specifically, it first predicts Gaussians in each local camera coordinate system which was transformed to the world coordinate using camera poses.
For fair comparisons, both methods use the same backbone and head but differ in the prediction of Gaussian space.
The results in row (f) of Tab.~\ref{tab:ablation} show that our pose-free canonical space prediction method outperforms such pose-required strategy. Fig.~\ref{fig:ablation_vis} illustrates that the transform-then-fuse strategy leads to the ghosting artifacts in the rendering, because it struggles to align two separate Gaussians of two input views when transformed to a global coordinate.

\begin{minipage}[!tb]{\textwidth}
\begin{minipage}{.65\textwidth}
        \fontsize{7}{9}\selectfont
        \centering
        \includeinkscape[width=1.0\linewidth]{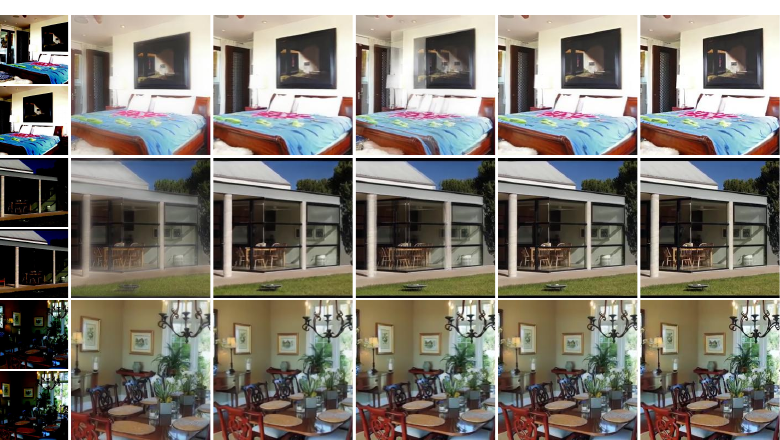}
        \captionof{figure}{\textbf{Ablations.} No intrinsic results in blurriness due to scale misalignment. Without the RGB image shortcut, the rendered images are blurry in the texture-rich areas. Using the transform-then-fuse strategy causes ghosting problem.}
        \label{fig:ablation_vis}
\end{minipage}\quad
\begin{minipage}{.32\textwidth}
        \centering
        \scalebox{0.55}{
        \begin{tabular}{llccc}
        \toprule
        Num & Design & PSNR & SSIM & LPIPS \\
        \midrule
        (a) & Ours & \best{25.033} & \best{0.838} & \best{0.160} \\
        \midrule
        (b) & No intrinsic emb. & 23.543 & 0.780 & 0.186 \\
        (c) & w/ dense emb. & 24.809 & 0.832 & 0.166 \\
        (d) & w/ global emb. - add. & 24.952 & 0.836 & 0.161 \\
        \midrule
        (e) & No RGB shortcut & 24.279 & 0.810 & 0.183 \\
        \midrule
        (f) & Transform-then-fuse & 24.632 & 0.834 & 0.167 \\
        \midrule
        (g) & \color{gray} 3 input views & \color{gray} 26.619 & \color{gray} 0.872 &  \color{gray} 0.127 \\
        \bottomrule
        \end{tabular}}
        \captionof{table}{\textbf{Ablations.} intrinsic embeddings are vital for performance and using intrinsic tokens performs the best. Adding the RGB image shortcut also improves the quality of rendered images. Our method achieves better performance compared with the pose-required per-local-view Gaussian field prediction method.}
        \label{tab:ablation}
\end{minipage}
\end{minipage}

\textbf{Ablation on Camera Intrinsic Embedding.}
Here we study three intrinsic encodings described in Sec.~\ref{subsec:intrinsic} as well as inputting no intrinsic information. First, we can see in Tab.~\ref{tab:ablation} row (b) and Fig.~\ref{fig:ablation_vis} that no intrinsic encodings lead to blurry results as the scale ambiguity confuses the learning process of the model. 
Second, we notice that the intrinsic token constantly performs the best marginally among these three proposed intrinsic encodings, so, we use it as our default intrinsic encoding.

\textbf{Importance of RGB Shortcut.}
As discussed in Sec.~\ref{sec:network}, in addition to the low-res ViT features, we also input RGB images into the Gaussian parameter prediction head.
As shown in Fig.~\ref{fig:ablation_vis}, when there is no RGB shortcut, the rendered images are blurry in the texture-rich areas, see the quilt in row 1 and chair in row 3.

\textbf{Extend to 3 Input Views.}
For fair comparisons with baselines, we primarily conduct experiments using two input view settings. Here we present results using three input views by adding an additional view between the two original input views. As shown in row (g) of Tab.~\ref{tab:ablation}, the performance significantly improves with the inclusion of the additional view.

\section{Conclusion}
This paper introduces a simple yet effective pose-free method for generalizable sparse-view 3D reconstruction. By predicting 3D Gaussians directly in a canonical space from any given unposed multi-view images, we demonstrate superior performance in novel view synthesis and relative pose estimation. 
While our method currently applies only to static scenes, extending our pipeline to dynamic scenarios presents an interesting direction for future work.

\bibliography{iclr2025_conference}
\bibliographystyle{iclr2025_conference}

\appendix
\clearpage
\appendix

\section{More Implementation Details}
\textbf{More Training Details.}
We first describe the process for training the $256\times256$ model, which serves as the basis for all baseline comparisons. Following \citet{mvsplat}, when training on RealEstate10K~\citep{re10k} and ACID~\citep{acid} separately, the model is trained on $2.4 \times 10^{6}$ input image pairs, randomly sampled from training video clips to ensure diverse input coverage. 
We employ the AdamW optimizer~\citep{adamw}, setting the initial learning rate for the backbone to $2\times10^{-5}$ and other parameters to $2\times10^{-4}$.
The model is trained using 8 GH200 GPUs with a total batch size of 128, which takes approximately 6 hours to complete.
When trained on the combined RealEstate10k and DL3DV~\citep{dl3dv} datasets, we sample image pairs evenly from both datasets and double the training steps.

We then train the $512\times512$ model using the pre-trained weights of the $256\times256$ model. The $512\times512$ model is also trained on the combined RealEstate10K and DL3DV datasets, following the same training procedure. Since no fair baseline model is available for comparison, we focus primarily on the qualitative results of the $512\times512$ model. For instance, the results presented in Fig.~\ref{fig:wild_test} are obtained using the $512\times512$ model.

\textbf{Evaluation Set Generation.}
For the evaluation of the novel view synthesis task, we randomly select one image pair from each video sequence in the official test splits of RealEstate10K~\citep{re10k} and ACID~\citep{acid}. We then randomly sample three frames between the input views as target views. For the pose estimation task, we use the same input image pair as in the novel view synthesis task, and employ the COLMAP pose as the ground truth.

To determine the overlap ratio of image pairs, we employ RoMa~\citep{roma}, a state-of-the-art dense feature matching method. The process for calculating the overlap ratio for an image pair $\left\{\boldsymbol{I}^1, \boldsymbol{I}^2\right\}$ is as follows:

\begin{enumerate}
    \item Obtain dense image matching from $\boldsymbol{I}^1$ to $\boldsymbol{I}^2$ and vice versa.
    \item Consider matching scores above 0.005 as valid.
    \item Calculate the overlap ratio $r_{i \rightarrow j}$ for image $\boldsymbol{I}^i$ to image $\boldsymbol{I}^j$ as:
        \begin{equation*}
            r_{i \rightarrow j} = \frac{\text{Number of valid matched pixels}}{\text{Total number of pixels}}
        \end{equation*}
    \item Compute $r_{1 \rightarrow 2}$ and $r_{2 \rightarrow 1}$.
    \item Define the final overlap ratio as:
        \begin{equation*}
            r_{\text{overlap}} = \min(r_{1 \rightarrow 2}, r_{2 \rightarrow 1})
        \end{equation*}
\end{enumerate}

The resulting evaluation set and the code used for its generation will be made publicly available to facilitate further research in this area.

\textbf{Statistics on the Evaluation Set.}
The number of scenes in RealEstate10K \citep{re10k} for each overlap category is as follows: 1403 for small, 2568 for medium, and 1630 for large overlaps. In the ACID dataset \citep{acid}, the number of scenes for each overlap category is: 249 for small, 644 for medium, and 448 for large overlaps.

\textbf{Baseline Setup for Pose Estimation Tasks.}
For a fair comparison with RoMa~\citep{roma}, we first resize and center-crop the input images to match the dimensions used in our model ($256 \times 256$), then resize them to a coarse resolution (560 × 560) and an upsampled resolution (864 × 864) to fit RoMa’s requirements. However, since DUSt3R~\citep{dust3r} and MASt3R~\citep{mast3r} are not trained with square image inputs, simply setting the input resolution to $256 \times 256$ leads to poor pose estimation performance. Therefore, we resize and center-crop their input images to $512 \times 256$, as officially supported by their models. Consequently, the image content visible to DUSt3R and MASt3R is greater than that seen by RoMa and our model.

\textbf{Details on Pose Estimation.}
After obtaining the coarse camera pose, we apply photometric loss-based optimization to refine it. This optimization is performed for 200 steps per input image pair with a learning rate of $5\times10^{-3}$, taking approximately 2 seconds. Notably, our pose estimation can also operate without the optimization refinement process, resulting in an acceptable performance degradation but decreasing inference time. As shown in Tab.~\ref{table:ablation_pose}, despite the performance drop, our method still performs comparably to RoMa, which is trained on ScanNet, even in a zero-shot setting.

\textbf{Details on Applying to In-the-Wild Data.}
Our method requires camera intrinsic information as input for rendering novel views. When testing our model on out-of-distribution datasets, such as Tanks and Temples~\citep{tnt}, we simply use the ground truth intrinsic data provided by the dataset. When tested on photos taken with mobile phones, we extract the focal length from their EXIF metadata. For data generated by SORA~\citep{sora}, we set the focal length to $(H + W) / 2$. We found that this heuristic setting works quite well, as our model is relatively robust to focal length variations after training.

\section{More Experimental Analysis}
\textbf{Ablations on Backbone Initialization.}
\begin{table}[t]
    \centering
    \caption{\textbf{Ablation on different weight initialization}.}
    \begin{adjustbox}{max width=\textwidth}
    \begin{tabular}{lccc ccc ccc ccc}
        \toprule
        & \multicolumn{3}{c}{Small} & \multicolumn{3}{c}{Medium} & \multicolumn{3}{c}{Large} & \multicolumn{3}{c}{Average} \\
        \cmidrule(lr){2-4} \cmidrule(lr){5-7} \cmidrule(lr){8-10} \cmidrule(lr){11-13}
        &  PSNR$\uparrow$ & SSIM$\uparrow$ & LPIPS$\downarrow$ & PSNR$\uparrow$ & SSIM$\uparrow$ & LPIPS$\downarrow$ & PSNR$\uparrow$ & SSIM$\uparrow$ & LPIPS$\downarrow$ & PSNR$\uparrow$ & SSIM$\uparrow$ & LPIPS$\downarrow$ \\
        \midrule
        MASt3R & \best{22.514} & \best{0.784} & \best{0.210} & \best{24.899} & \best{0.839} & \best{0.160} & \best{27.411} & \best{0.883} & \best{0.119} & \best{25.033} & \best{0.838} & \best{0.160} \\
        DUSt3R & 22.281 & 0.772 & 0.217 & 24.444 & 0.824 & 0.168 & 26.679 & 0.865 & 0.128 & 24.553 & 0.823 & 0.169 \\
        CroCoV2 & 22.181 & 0.761 & 0.222 & 24.423 & 0.819 & 0.170 & 26.821 & 0.866 & 0.128 & 24.559 & 0.818 & 0.171 \\
        \bottomrule
    \end{tabular}
    \end{adjustbox}
\label{tab:ablation_weight_init}
\end{table}

We initialize our backbone network with MASt3R~\citep{mast3r} pre-trained weights in our main experiences. MASt3R is trained on datasets with ground truth depth annotation, but notably, its training data has no overlap with the training and evaluation datasets used in our experiences. 
Here, we also explore the effect of different backbone initialization methods. We compare the performance of our method with DUSt3R~\citep{dust3r} and CroCo V2~\citep{crocov2} pre-trained weights separately. 
DUSt3R is also trained with depth supervision but without feature matching loss compared with MASt3R. CroCoV2 is pre-trained with pure 2D image pairs with reconstruction loss~\citep{mae}. Note that for CroCoV2 initialization, we warm up the training by adding point cloud distillation loss from the DUSt3R model for 1,000 steps, this aims to tell the network the goal is to predict the Gaussians in the canonical space, otherwise, the learning target is too hard for the network to understand as we only train our network with photometric loss.
The results in Tab.~\ref{tab:ablation_weight_init} demonstrate that our method, even when initialized without depth-supervised pre-trained weights, performs similarly to models initialized with DUSt3R and MASt3R. Moreover, it consistently outperforms previous state-of-the-art pose-required methods (pixelSplat\citep{pixelsplat} and MVSplat~\citep{mvsplat}) on scenes with small overlap, while offering comparable performance on scenes with large overlap.
This indicates that our canonical Gaussian space representation can be learned without depth pre-trained weights. 
However, we also find that using MASt3R pre-trained weights performs slightly better than others. This improvement may brought by the feature matching loss used in MASt3R, and we will leave this as a further exploration direction.

\textbf{Ablations on the Effectiveness of Two-Stage Pose Estimation.}
\begin{table}[t]
\centering
\caption{\textbf{Ablation on the pose estimation method}.}
\begin{adjustbox}{max width=\textwidth}
\begin{tabular}{ccccc}
\toprule
PnP & Photometric & 5$^\circ \uparrow$     & 10$^\circ \uparrow$    & 20$^\circ \uparrow$    \\ 
\midrule
\checkmark   & \checkmark          & \best{0.318} & \best{0.538} & \best{0.717} \\
\checkmark   &             & 0.287 & 0.506 & 0.692 \\
    & \checkmark           & 0.017 & 0.027 & 0.051 \\
\bottomrule
\end{tabular}
\end{adjustbox}
\label{table:ablation_pose}
\end{table}

As shown in Tab.~\ref{table:ablation_pose}, relying solely on PnP-RANSAC results in inaccurate pose estimates. However, if we skip the coarse camera pose estimation stage and only use photometric loss-based optimization starting from $[\boldsymbol{I}\mid\boldsymbol{0}]$, the performance significantly degrades when using the same number of optimization steps (200 steps). This is because optimizing from an initial pose far from the target is more challenging.

\textbf{Evaluate on the Evaluation Set of pixelSplat.}
\begin{table}[t]
\centering
\caption{\textbf{Performance comparison on the evaluation set of pixelSplat.}}
\begin{tabular}{lccc}
\toprule
& PSNR$\uparrow$ & SSIM$\uparrow$ & LPIPS$\downarrow$ \\
\midrule
pixelNeRF & 20.43 & 0.589 & 0.55 \\
GPNR & 24.11 & 0.793 & 0.255 \\
AttnRend & 24.78 & 0.82 & 0.213 \\
pixelSplat & 26.09 & 0.863 & 0.136 \\
MVSplat & 26.39 & 0.869 & 0.128 \\
Ours & \best{26.786} & \best{0.878}  & \best{0.124} \\
\bottomrule
\end{tabular}
\label{tab:re10k_old}
\end{table}

We also present results based on the evaluation set used by pixelSplat~\citep{pixelsplat} and MVSplat~\citep{mvsplat}. The results are shown in Tab.~\ref{tab:re10k_old}. However, we do not prioritize this evaluation set as it is relatively simple, with most input pairs exhibiting significant overlap, making it less suitable for more advanced future research.

\begin{figure*}[t]
    \centering
    \fontsize{9}{11}\selectfont
    \includeinkscape[width=0.7\linewidth]{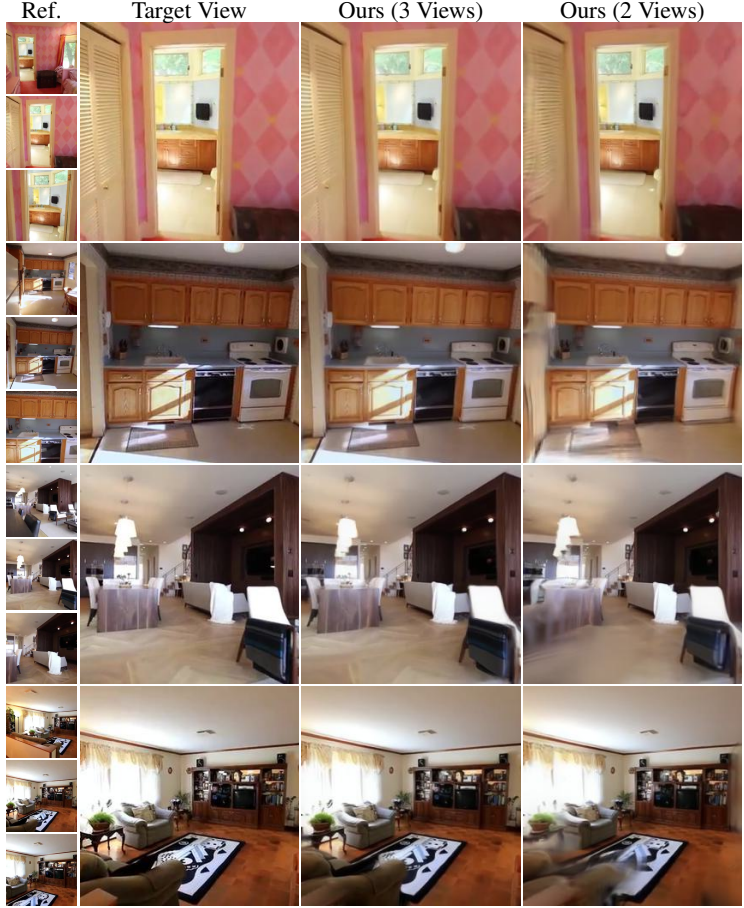}
    \caption{\textbf{Qualitative comparison on different numbers of input views.} The model with 2 input views utilizes only the top and bottom reference images, whereas the 3-view model incorporates an additional intermediate view. Adding this extra reference view improves the quality of rendered novel views significantly, as it captures finer spatial details and reduces occlusion effects.
    }
    \label{fig:vis_3views}
\end{figure*}

\textbf{Details on Extension to 3 Input Views.}
Our method can be extended to an arbitrary number of input views. As shown in Tab.~\ref{tab:ablation}, using three input views improves the results. Specifically, we add the middle frame between the original two input frames as an additional view. For three-input settings, when generating the Gaussians of each view, we concatenate the feature tokens from all other views and perform cross-attention with these concatenated tokens in the ViT decoder stage, keeping all other operations the same as with two input views. Fig.~\ref{fig:vis_3views} provides a qualitative comparison between using two and three input views.

\begin{table}[t]
    \centering
    \caption{\textbf{Performance of retrained Splatt3R~\citep{splatt3r} model.}}
    \begin{adjustbox}{max width=\textwidth}
    \begin{tabular}{lccc ccc ccc ccc}
        \toprule
        & \multicolumn{3}{c}{Small} & \multicolumn{3}{c}{Medium} & \multicolumn{3}{c}{Large} & \multicolumn{3}{c}{Average} \\
        \cmidrule(lr){2-4} \cmidrule(lr){5-7} \cmidrule(lr){8-10} \cmidrule(lr){11-13}
        &  PSNR$\uparrow$ & SSIM$\uparrow$ & LPIPS$\downarrow$ & PSNR$\uparrow$ & SSIM$\uparrow$ & LPIPS$\downarrow$ & PSNR$\uparrow$ & SSIM$\uparrow$ & LPIPS$\downarrow$ & PSNR$\uparrow$ & SSIM$\uparrow$ & LPIPS$\downarrow$ \\
        \midrule
        Ours & \best{22.514} & \best{0.784} & \best{0.210} & \best{24.899} & \best{0.839} & \best{0.160} & \best{27.411} & \best{0.883} & \best{0.119} & \best{25.033} & \best{0.838} & \best{0.160} \\
        Splatt3R Official & 14.352 & 0.475 & 0.472 & 15.529 & 0.502 & 0.425 & 15.817 & 0.483 & 0.421 & 15.318 & 0.490 & 0.436 \\
        Splatt3R Retrain & 17.987 & 0.616 & 0.385 & 19.362 & 0.657 & 0.327 & 20.518 & 0.685 & 0.288 & 19.354 & 0.655 & 0.330 \\
        \bottomrule
    \end{tabular}
    \end{adjustbox}
\label{tab:splatt_retrain}
\end{table}

\textbf{Addtional Comparison with Splatt3R.}
In Tab.\ref{tab:re10k} of the main paper, we compare our method with the official model provided by the authors of Splatt3R\citep{splatt3r}, which freezes the MASt3R model and is then trained on the ScanNet++~\citep{scannetpp} dataset.
Here, we also train it on the RealEstate10K~\citep{re10k} dataset at the same resolution as other baselines ($256 \times 256$). However, when attempting to retrain it using the official code provided by the authors on RealEstate10k, we find that the training fails because the original Splatt3R can only be trained on datasets with metric pose ground truth. This limitation arises because Splatt3R relies on a fixed pre-trained MASt3R~\citep{mast3r} model and its capability for metric depth prediction. As MASt3R is trained on ScanNet++~\citep{scannetpp} using ground truth metric depth information, the poses in ScanNet++ are also metric. Consequently, the original Splatt3r cannot be trained on video datasets without metric pose information. We identify two main issues:
\begin{enumerate}
    \item The camera poses provided by RealEstate10K are up-to-scale, as they are estimated using COLMAP~\citep{colmap}. This results in a scale misalignment between the ground truth pose and the scale of the estimated Gaussian field. Although Splatt3r predicts additional point offsets based on the point cloud estimated by MASt3R as the final Gaussian center, these offsets are typically small and unable to resolve the misalignment problem. This issue persists even in datasets with metric camera poses, as the point cloud provided by MASt3R is imperfect and not fully aligned with the ground truth pose scale. The training fails if the provided COLMAP target camera poses are used to render target views for the 3D Gaussians. To address this, we first estimate the camera poses of two input views using the point cloud from MASt3R, which is scale-consistent with the Gaussians. We then align the scale of the COLMAP target poses with the scale of the MASt3R point cloud by adjusting the scale of the input view poses to match the MASt3R-estimated ones.
    \item We also find that the intrinsic parameters estimated by MASt3R~\citep{mast3r} do not align well with the ground truth intrinsic parameters. Using the ground truth intrinsic during training also causes failure. As a result, we opt to use the intrinsics estimated by MASt3R.
\end{enumerate}
Additionally, we ignore the loss mask used in the original Splatt3R as it requires ground truth depth to generate the mask, but the RealEstate10K dataset lacks ground truth depth. The mask is unnecessary for training on video datasets like RealEstate10K, since during training, we use intermediate frames between the two input frames as targets, and most of the image content in the target frames is well covered by the input frames.
For a fair comparison during evaluation, we apply the same evaluation-time pose alignment technique used in our method. The results are presented in Tab.~\ref{tab:splatt_retrain}. Although retraining the model on the RealEstate10K dataset yields improved performance, it still significantly lags behind our approach. This performance gap can be attributed to the misalignment issues inherent in the fixed MASt3R model and the persistent scale ambiguity problem.

\section{Limitations}
Our approach, like previous pose-free methods \citep{coponerf, colmap-free3DGS}, assumes known camera intrinsics. Although heuristically set intrinsic parameters prove effective for in-the-wild images, relaxing this requirement would enhance the robustness of real-world applications. 
Additionally, as our feedforward model is non-generative, it lacks the ability to reconstruct unseen regions of a scene with detailed geometry and texture, as demonstrated in the results of the 2-view model shown in Fig.~\ref{fig:vis_3views}. This limitation could potentially be mitigated by incorporating additional input views, which would enhance scene coverage.
Finally, the current training data (limited to RealEstate10K, ACID, and DL3DV) constrains the model's generalizability to diverse in-the-wild scenarios. Future work could explore training our model on large-scale, diverse indoor and outdoor datasets, leveraging our method's independence from ground-truth depth information during training.

\section{More Visual Comparisons}
\begin{figure*}[t]
    \centering
    \fontsize{9}{11}\selectfont
    \includeinkscape[width=1.1\linewidth]{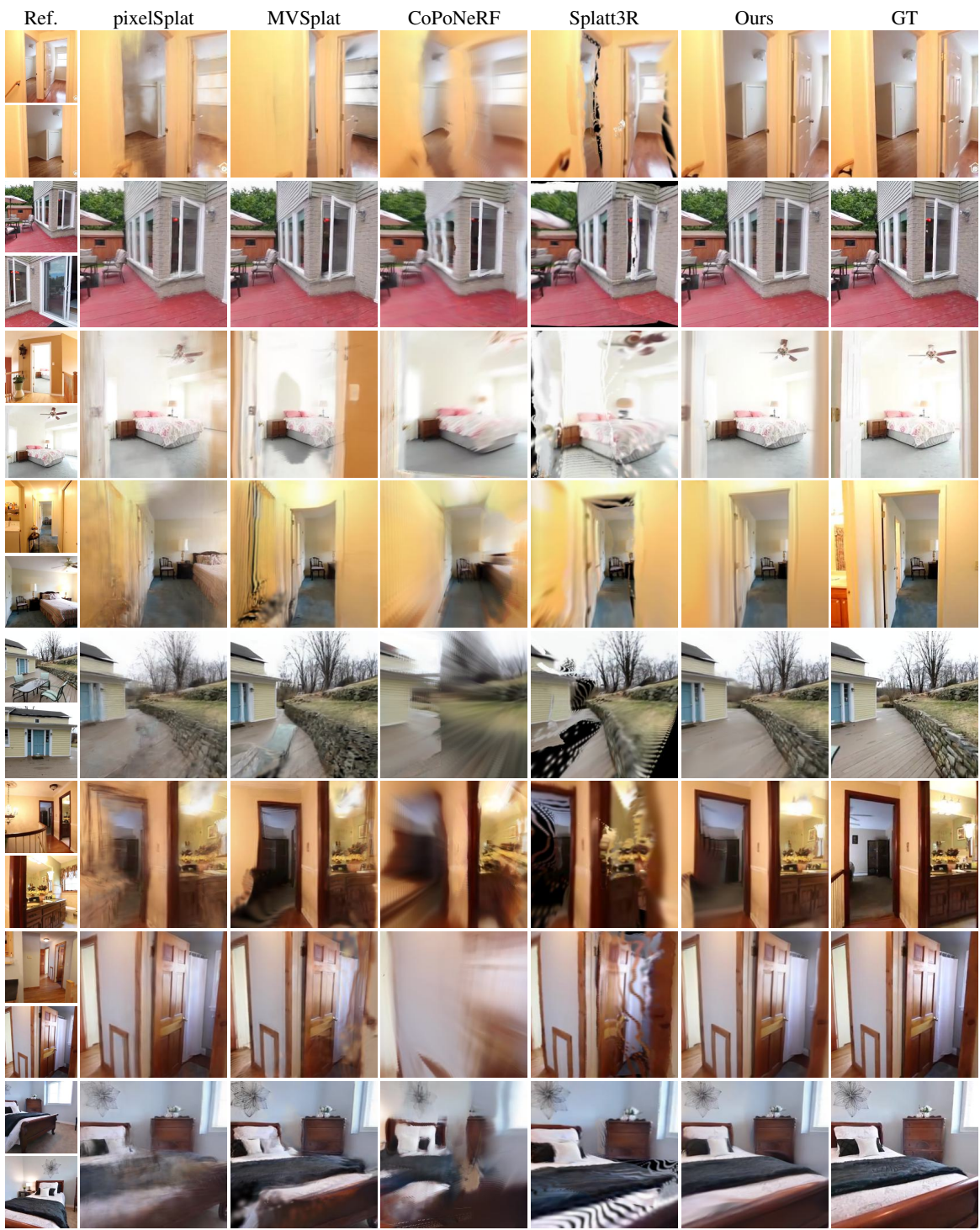}
    \caption{\textbf{More comparisons of the RealEstate10K dataset with \textit{small} overlap of input images.} 
    }
    \label{fig:re10k_small_comparison}
\end{figure*}

\begin{figure*}[t]
    \centering
    \fontsize{9}{11}\selectfont
    \includeinkscape[width=1.1\linewidth]{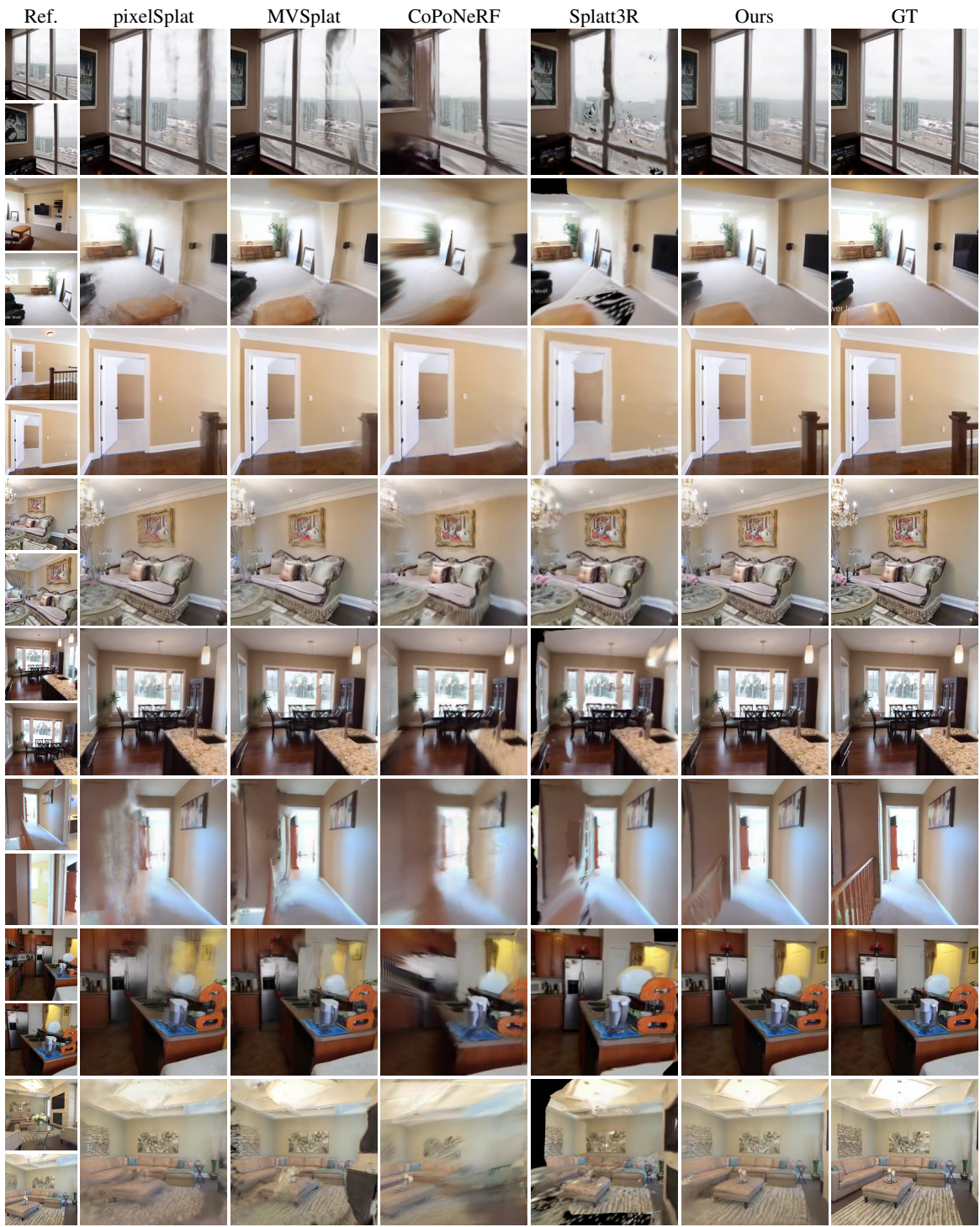}
    \caption{\textbf{More comparisons of the RealEstate10K dataset with \textit{medium} overlap of input images.} 
    }
    \label{fig:re10k_medium_comparison}
\end{figure*}

\begin{figure*}[t]
    \centering
    \fontsize{9}{11}\selectfont
    \includeinkscape[width=1.1\linewidth]{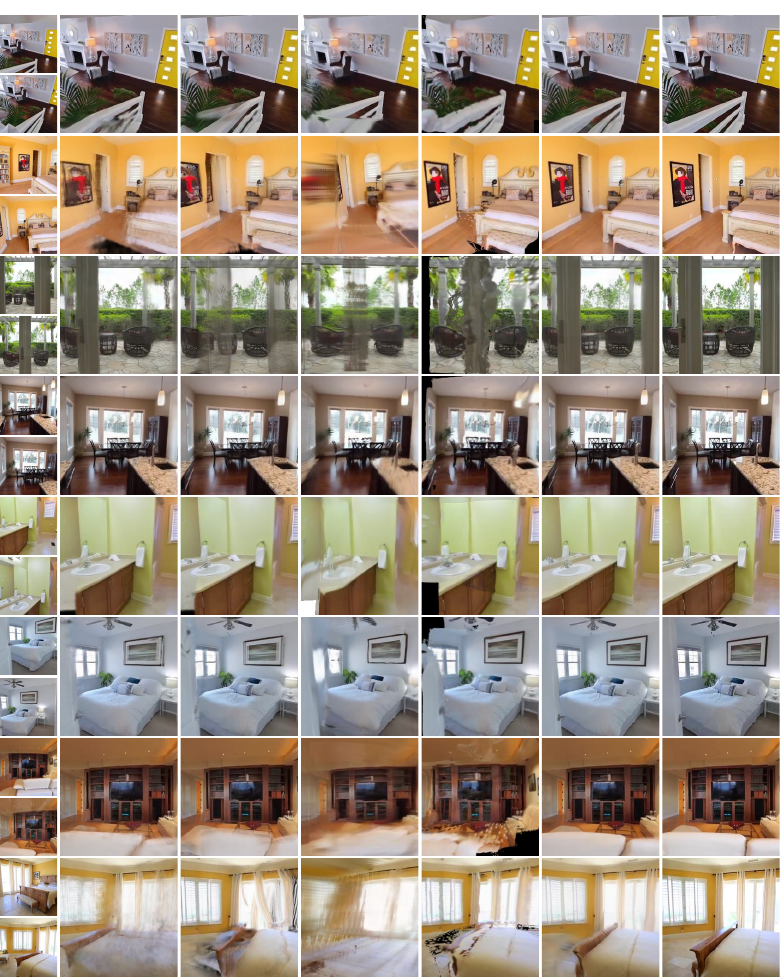}
    \caption{\textbf{More comparisons of the RealEstate10K dataset with \textit{large} overlap of input images.} 
    }
    \label{fig:re10k_large_comparison}
\end{figure*}

\begin{figure*}[t]
    \centering
    \fontsize{9}{11}\selectfont
    \includeinkscape[width=1.1\linewidth]{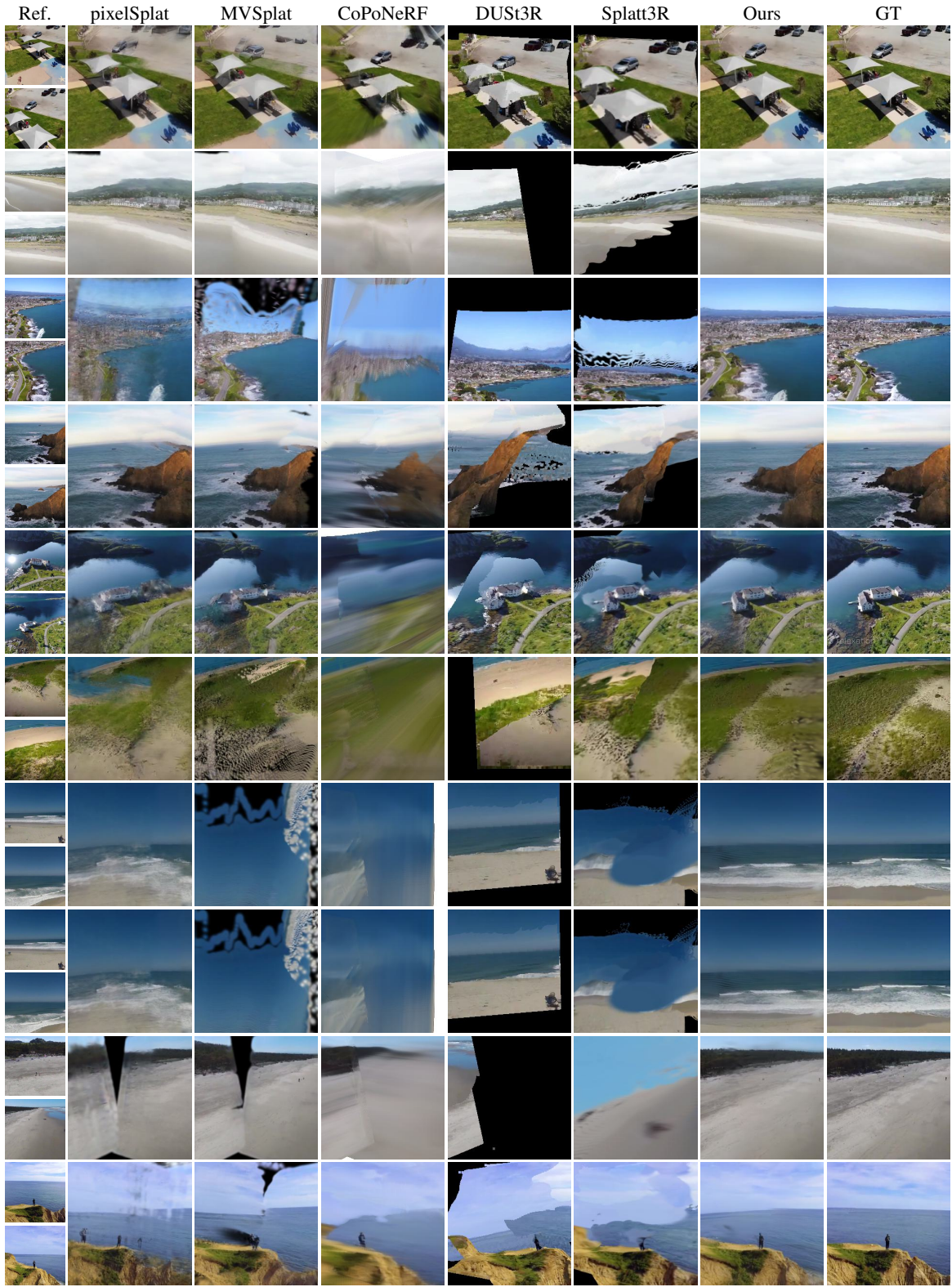}
    \caption{\textbf{More comparisons on the ACID dataset.} 
    }
    \label{fig:acid_comparison}
\end{figure*}

We present an additional comparison with previous SOTA pose-dependent and pose-free methods across various levels of image overlap: Fig.\ref{fig:re10k_small_comparison} for small input image overlap, Fig.\ref{fig:re10k_medium_comparison} for medium input image overlap, and Fig.\ref{fig:re10k_large_comparison} for large input image overlap. Furthermore, we provide additional comparisons on the ACID dataset \citep{acid} in Fig.\ref{fig:acid_comparison}.

\end{document}